\documentclass[preprint,11pt]{elsarticle}
\pdfoutput=1
\usepackage{amssymb}
\usepackage{enumitem}
\usepackage{geometry}
\usepackage{amsmath}
\usepackage{diagbox}
\usepackage{color}
\usepackage{multirow}
\usepackage{booktabs}

\begin{document}

\begin{frontmatter}

\title{Improving short-term bike sharing demand forecast through an irregular convolutional neural network}

\author[label1]{Xinyu Li}
\author[label1,label2]{Yang Xu \corref{cor1}}
\cortext[cor1]{Corresponding author: yang.ls.xu@polyu.edu.hk}
\author[label3]{Xiaohu Zhang}
\author[label1,label4]{Wenzhong Shi}
\author[label5]{Yang Yue}
\author[label5]{Qingquan Li}

\address[label1]{Department of Land Surveying and Geo-Informatics, The Hong Kong Polytechnic University}
            
\address[label2]{The Hong Kong Polytechnic University Shenzhen Research Institute}
             
\address[label3]{Department of Urban Planning and Design, The University of Hong Kong}
             
\address[label4]{Smart Cities Research Institute, The Hong Kong Polytechnic University}

\address[label5]{Department of Urban Informatics, School of Architecture and Urban Planning, Shenzhen University} 

\begin{abstract}
As an important task for the management of bike sharing systems, accurate forecast of travel demand could facilitate dispatch and relocation of bicycles to improve user satisfaction. In recent years, many deep learning algorithms have been introduced to improve bicycle usage forecast. A typical practice is to integrate convolutional (CNN) and recurrent neural network (RNN) to capture spatial-temporal dependency in historical travel demand. For typical CNN, the convolution operation is conducted through a kernel that moves across a ``matrix-format'' city to extract features over spatially adjacent urban areas. This practice assumes that areas close to each other could provide useful information that improves prediction accuracy. However, bicycle usage in neighboring areas might not always be similar, given spatial variations in built environment characteristics and travel behavior that affect cycling activities. Yet, areas that are far apart can be relatively more similar in temporal usage patterns. To utilize the hidden linkage among these distant urban areas, the study proposes an irregular convolutional Long-Short Term Memory model (IrConv+LSTM) to improve short-term bike sharing demand forecast. The model modifies traditional CNN with irregular convolutional architecture to extract dependency among ``semantic neighbors''. The proposed model is evaluated with a set of benchmark models in five study sites, which include one dockless bike sharing system in Singapore, and four station-based systems in Chicago, Washington, D.C., New York, and London. We find that IrConv+LSTM outperforms other benchmark models in the five cities. The model also achieves superior performance in areas with varying levels of bicycle usage and during peak periods. The findings suggest that ``thinking beyond spatial neighbors'' can further improve short-term travel demand prediction of urban bike sharing systems.
\end{abstract}

\begin{keyword}
bike sharing \sep deep learning \sep travel demand forecast \sep spatial-temporal analysis \sep irregular convolution
\end{keyword}

\end{frontmatter}

\section{Introduction}
Shared bicycles have received increasing attention in urban transportation during the past few decades~\cite{midgley2011bicycle,larsen2013bike}. As a green transport option for short-distance travel in cities, bike-sharing services can reduce carbon emissions and enhance last-mile connectivity to public transit~\cite{litman2006issues, steg2005sustainable, haghshenas2012urban}. During COVID-19 pandemic, bike-sharing is found to be a more resilient mode that can mitigate the fear of overcrowding in public transit~\cite{jobe2021bike, hu2021examining,kim2021impact}. Given the importance of bike-sharing services in urban transportation, accurate demand forecasting is crucial for effective rebalancing in daily operations. Many studies attempted to develop frameworks that can accurately estimate the bicycle demand throughout the city by applying traditional and machine learning models~\cite{raviv2013optimal, dell2018bike,dell2014bike, singhvi2015predicting, kumar1999autoregressive, avuglah2014application, billings2006application, lee1999application}

In recent years, deep learning approaches have been widely used for predicting short-term traffic demand~\cite{ai2019deep, pan2019predicting, li2017diffusion, zhang2017deep, li2021short, zhang2020novel, ren2020hybrid,sathishkumar2020using,yang2020using}. A critical task is to model the spatial-temporal dependency in travel demand. Two mainstream architectures, Convolutional Neural Network (CNN) and Recurrent Neural Network (RNN), are often integrated to capture the spatial and temporal information of traffic demand~\cite{ma2017learning,xiangxue2019data, fu2016using,zhao2017lstm,du2019deep}. Typically, CNN exploits regular convolutional kernels scanning through the input features (e.g., images) to extract spatial characteristics of travel demand~\cite{lecun2015deep}. RNN leverages the extracted temporal dynamic behavior from the past elements of the sequence to predict the next element~\cite{lecun2015deep,hochreiter1997long}. To better capture spatial-temporal information, several hybrid deep learning frameworks incorporating both CNN and RNN architectures are developed and these models achieved good performance in various traffic prediction tasks~\cite{ai2019deep, li2021short, zhang2020novel,ren2020hybrid}. 

However, CNN has certain shortcomings when it is employed to capture spatial-temporal information of bike sharing demand. CNN achieves desirable performance in object detection for images, because adjacent pixels of the same object are often highly correlated. Unlike images, bicycle usage in neighboring urban areas can be quite different due to spatial variations of travel behavior and built environmental characteristics~\cite{du2019model,xu2019unravel}. On the other hand, for certain areas that are far apart, the bicycle usage patterns may exhibit similar temporal rhythms. Given the regular shape of convolutional kernels, deep learning models with typical CNN architecture are not able to capture the similarities of bike usage patterns among distant urban areas. If such similarities can be captured and incorporated into the prediction models, it may further enhance the accuracy and reliability of bike sharing demand forecast.

To bridge the research gap, this paper introduces an irregular convolutional Long Short-Term Memory model (IrConv+LSTM) to improve short-term demand forecast for urban bike-sharing systems. The model employs irregular convolutional architecture to capture the dependency of bicycle usage among distant urban areas. Given the areas being forecast, an irregular convolution operation is performed over their \textit{semantic neighbors}, which refer to places that show similar temporal bicycle usage patterns. Two measures, namely Pearson Correlation Coefficient (IrConv+LSTM:P) and Dynamic Time Warping (IrConv+LSTM:D), are used as similarity metrics to identify semantic neighbors for the areas being forecast. The two variants of the proposed model (IrConv+LSTM:P and IrConv+LSTM:D) and several benchmark models are evaluated and compared over bike-sharing systems in five cities, including one dockless bike-sharing system in Singapore and four station-based systems in Washington D.C., Chicago, New York, and London, respectively.

The remainder of this paper is organized as follows. Section~\ref{review} provides a review of relevant literature. Study sites and related terminologies are introduced in Section~\ref{study_area}. Section~\ref{method} and \ref{results} illustrate the methodologies and experimental results. In Section~\ref{concolusion}, we conclude the study and discuss possible future research directions.

\section{Literature Review}\label{review}
Many studies have been conducted on traffic demand prediction using either parametric or non-parametric models~\cite{nagy2018survey}. In parametric models, data series is modelled as a dynamic variation from a systemic basis. Most parametric models adopt filters to estimate parameters that capture the system characteristics for predicting future status. Autoregressive Integrated Moving Average (ARIMA) model and Kalman Filter are two typical parametric models~\cite{xu2017real}. ARIMA model uses Autoregressive (AR) or Moving Average(MA) methods to simulate the temporal autocorrelation of a smoothed sequence. The applications of ARIMA in traffic prediction include forecasting traffic accidents, traffic status from the perspective of speed, volume and travel time~\cite{kumar1999autoregressive, avuglah2014application, billings2006application,lee1999application}. Kalman Filtering is an algorithm for optimal system state estimation by using status equations of linear systems. Similar to ARIMA, Kalman Filtering not only predicts traffic status but also assists traffic management and controls~\cite{antoniou2007nonlinear, szeto1972application, van2012applications}. Such parametric models filter a wealth of information based on the strong assumptions of the data. Also, spatial information is not considered in such models. Therefore, the parametric models cannot fully model the spatial-temporal characteristics from the historical traffic data. 

With the advancement of computing power, machine learning algorithms have been increasingly adopted for travel demand forecast.These algorithms include Support Vector Machine (SVM), Random Forest, Bayesian Network, Markov Model, Neural Network, and hybrid deep learning models~\cite{zhang2009traffic, hou2017road, sun2006bayesian,sun2005traffic}. In particular, deep learning models have been attracting much attention in the last decade. Many models are applied to predict the traffic status, such as Convolutional Neural Network (CNN), Recurrent Neural Network (RNN) with its variants, encoder-decoder and attention mechanisms for sequential prediction, and Graph Convolutional Network (GCN) for graphic knowledge learning~\cite{ma2017learning, zhao2017lstm, wang2020long, zi2021tagcn,liu2018short,zhang2018citywide}. For example, Ma \textit{et al.} establishes a space-time matrix representing traffic sensors ordered by road directions and time, then adopts CNN to capture spatial and temporal information for predicting~\cite{ma2017learning}. In \cite{fu2016using,zhao2017lstm,shin2020prediction}, RNN and LSTM are employed to predict the traffic status, including traffic speed and congestion. To incorporate spatial and temporal characteristics, many scholars also propose several hybrid models to forecast the traffic demands or flows. Zhang \textit{et al.} employs three residual CNNs to capture spatial-temporal information of historical pedestrian flows for predicting citywide crowd flows~\cite{zhang2017deep}. In~\cite{9136910}, CNN and attention LSTM are adopted to forecast passenger flow in urban rail transit. Ren \textit{et al.} adopts residual CNN and LSTM blocks for extracting high-level spatial-temporal information of pedestrian flow volumes to forecast citywide pedestrian volumes~\cite{ren2020hybrid}. In general, deep learning models has achieved better performance in predicting traffic status than the parametric models.

There are several studies for predicting bike-sharing demand using deep learning approaches. Several studies adopt GCN to capture spatial characteristics and employ attention mechanisms or fully connected networks to extract sequential information in SBSS for predicting bike-sharing demand~\cite{zi2021tagcn, kim2019graph, lin2018predicting}. Ai \textit{et al.} adopts convolutional LSTM to predict the bicycle demand in DBSS~\cite{ai2019deep}. However, the characteristics of bike-sharing systems in cities are not considered in the models, affecting the models' performance in predicting bike-sharing demand. Besides, few studies tested the proposed deep learning model in different city contexts to predict both SBSS and DBSS. Such types of cross-city and cross-system studies are essential to examine the generalizability of proposed deep learning architectures.

\section{Study Areas and Data Preprocessing}\label{study_area}
In this study, we use bike-sharing datasets in five different cities to assess the performance of our proposed model. These datasets include one dockless bike-sharing system (DBSS) in Singapore, and four station-based systems (SBSS) in Chicago, Washington D.C., New York, and London, respectively. The datasets of SBSS are collected from the bicycle trip records that document the departure and arrival time and stations. The dataset of DBSS in Singapore is collected from raw GPS coordinates of starting and ending locations of a trip. The GPS trajectories are preprocessed to remove GPS drifts and outliers using the approach from a prior study in Singapore~\cite{xu2019unravel}.

\begin{table}[htbp]
\caption{Description of Research Areas}
\centering
\setlength\extrarowheight{0.1pt}
\resizebox{\textwidth}{!}{
\begin{tabular}{cccccc} 
\bottomrule
\multicolumn{1}{l}{\diagbox{\textbf{Para.}}{\textbf{City}}}                                     & \textbf{Singapore}                                         & \textbf{Chicago}                                           & \begin{tabular}[c]{@{}c@{}}\textcolor[rgb]{0.2,0.2,0.2}{\textbf{Washington }}\\\textcolor[rgb]{0.2,0.2,0.2}{\textbf{D.C.}}\end{tabular} & \begin{tabular}[c]{@{}c@{}}\textbf{New York}\end{tabular} & \textbf{London}                                             \\ 
\bottomrule
\begin{tabular}[c]{@{}c@{}}\textbf{Spatial }\\\textbf{Resolution}\end{tabular}  & \multicolumn{5}{c}{1~Kilometre}                                                                                                                                                                                                                                                                                                                                                                           \\ 
\hline
\begin{tabular}[c]{@{}c@{}}\textbf{Temporal~}\\\textbf{Resolution}\end{tabular} & \multicolumn{5}{c}{1 Hour}                                                                                                                                                                                                                                                                                                                                                                                \\ 
\hline
\begin{tabular}[c]{@{}c@{}}\textbf{Training }\\\textbf{Period}\end{tabular}     & \begin{tabular}[c]{@{}c@{}}16/06-02/08,\\2017\end{tabular} & \begin{tabular}[c]{@{}c@{}}01/06-30/09,\\2019\end{tabular} & \begin{tabular}[c]{@{}c@{}}01/06-30/09,\\2019\end{tabular}                                                                              & \begin{tabular}[c]{@{}c@{}}01/06-30/09,\\2019\end{tabular}               & \begin{tabular}[c]{@{}c@{}}01/06-30/09,\\2019\end{tabular}  \\ 
\hline
\begin{tabular}[c]{@{}c@{}}\textbf{Validation}\\\textbf{Period}\end{tabular}    & \begin{tabular}[c]{@{}c@{}}03/08-31/08,\\2017\end{tabular} & \begin{tabular}[c]{@{}c@{}}01/10-25/10,\\2019\end{tabular} & \begin{tabular}[c]{@{}c@{}}01/10-31/10,\\2019\end{tabular}                                                                              & \begin{tabular}[c]{@{}c@{}}01/10-25/10,\\2019\end{tabular}               & \begin{tabular}[c]{@{}c@{}}01/10-31/10,\\2019\end{tabular}  \\
\bottomrule
\end{tabular}}
\label{table:study_area}
\end{table}
\begin{figure}[!h]
\centering
	\includegraphics[width=5.5in]{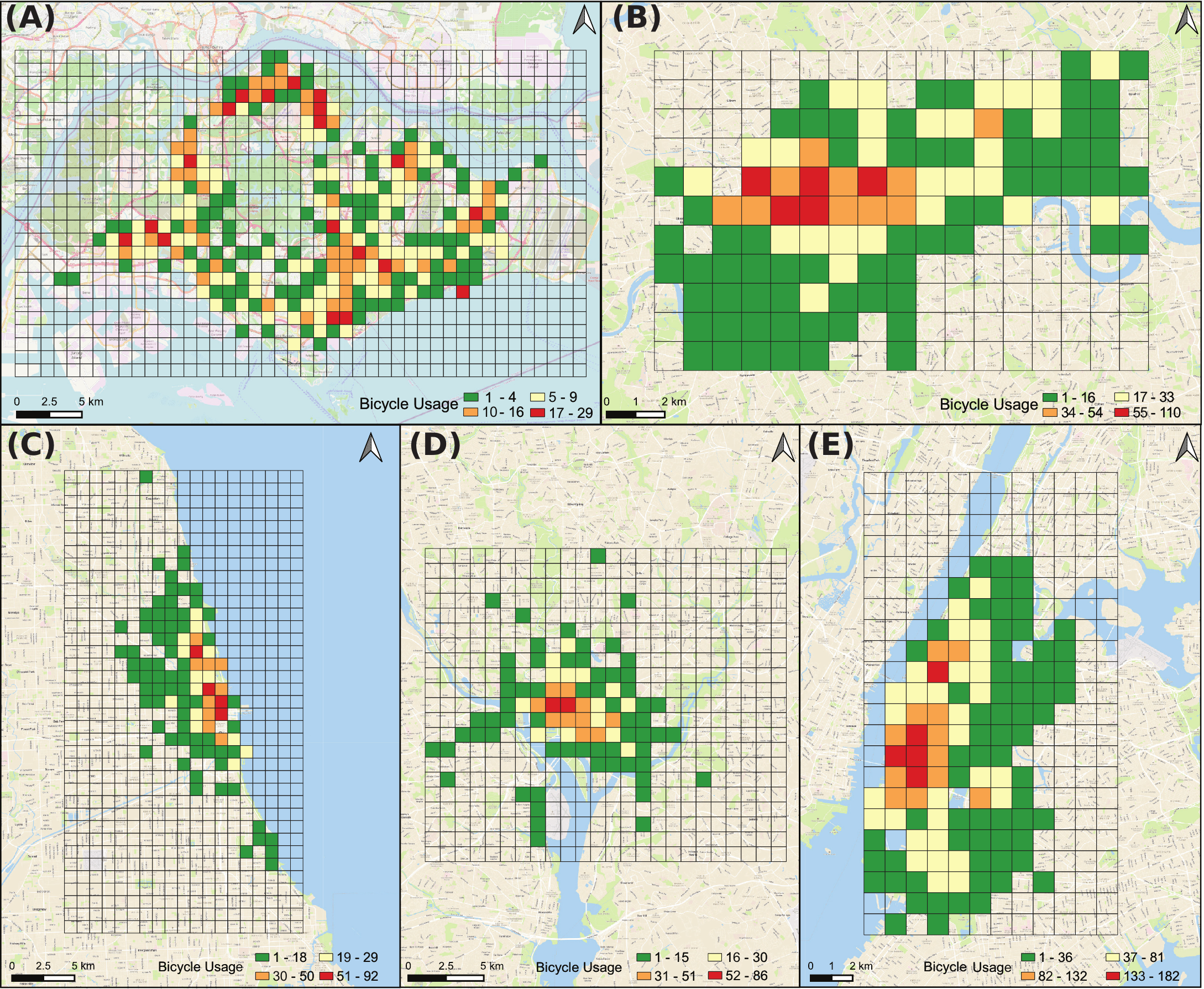}
	\caption{Spatial distribution of bicycle usage in five cities. \textbf{(A)} Singapore during 5-6 PM on August $25^{th}$, 2017; \textbf{(B)} London during 5-6 PM on August $24^{th}$, 2019; \textbf{(C)} Chicago during 1-2 PM on August $25^{th}$, 2019; \textbf{(D)} Washington, D.C. during 5-6 PM on August $25^{th}$, 2019; \textbf{(E)} New York during 5-6 PM on August $25^{th}$, 2019.}
	\label{fig:figure_grid_based}
\end{figure}
As shown in Fig.~\ref{fig:figure_grid_based}, we use regular grids to summarize the usage pattern of shared bicycles in the five cities for both DBSS and SBSS. A city area is divided into a regular $w*h$ grid map under a specific spatial resolution. During a time interval, the grid values represent the number of bicycle pick-ups distributed in the city. As shown in Table~\ref{table:study_area}, we set 1$km$ as the spatial resolution and 1 hour as the temporal resolution for this study. At the $k^{th}$ time interval, the definition of a grid map ($X_k(w,h)$) is shown in Eq.~\ref{equ:overall_demand}. $x_{k}(i,j)$ denotes the number of pick-ups in the cell located in the $i^{th}$ row and the $j^{th}$ column at the $k^{th}$ time interval, defined in Eq.~\ref{equ:cell_demand}. 
\begin{equation}
\resizebox{0.55\hsize}{!}{$
X_k(w,h) =  \left[
\begin{matrix}
{x_{k}(1,1)}&{x_{k}(1,2)}&{\cdots}&{x_{k}(1,h-1)}&{x_{k}(1,h)}\\
{x_{k}(2,1)}&{x_{k}(2,2)}&{\cdots}&{x_{k}(2,h-1)}&{x_{k}(2,h)}\\
{\vdots}&{\vdots}& {\ddots}&{\vdots}&{\vdots}\\
{x_{k}(w,1)}&{x_{k}(w,2)}&{\cdots}&{x_{k}(w,h-1)}&{x_{k}(w,h)}\\
\end{matrix}
\right]$}
\label{equ:overall_demand}
\end{equation}

\begin{equation}
x_{k}(i,j)= \mid \left \{T \in \mathbb{T}_{k} \mid T(O) \in (i,j)\wedge T(D) \notin (i,j) \right \}\mid
\label{equ:cell_demand}
\end{equation}

Here $\mathbb{T}_{k}$ denotes the trajectories occurred in $k^{th}$ time slot; $T(O)$ and $T(D)$ represent the departure and arrival locations of a trajectory $T$, respectively; $T(O) \in (i,j)\wedge T(D) \notin (i,j)$ denotes a trajectory starts from the cell $(i,j)$ but ends in another cell except the cell $(i,j)$; $\left | \cdot  \right |$ denotes the cardinality of a set. In this study, trips that start and end in the same cell are excluded because they do not affect the overall balance of bicycle supply and demand within a cell.

\section{Methodology}\label{method}
\subsection{Overall architecture of the proposed model}\label{architecture}
\begin{figure}[h!]
\centering
	\includegraphics[width=5.8in]{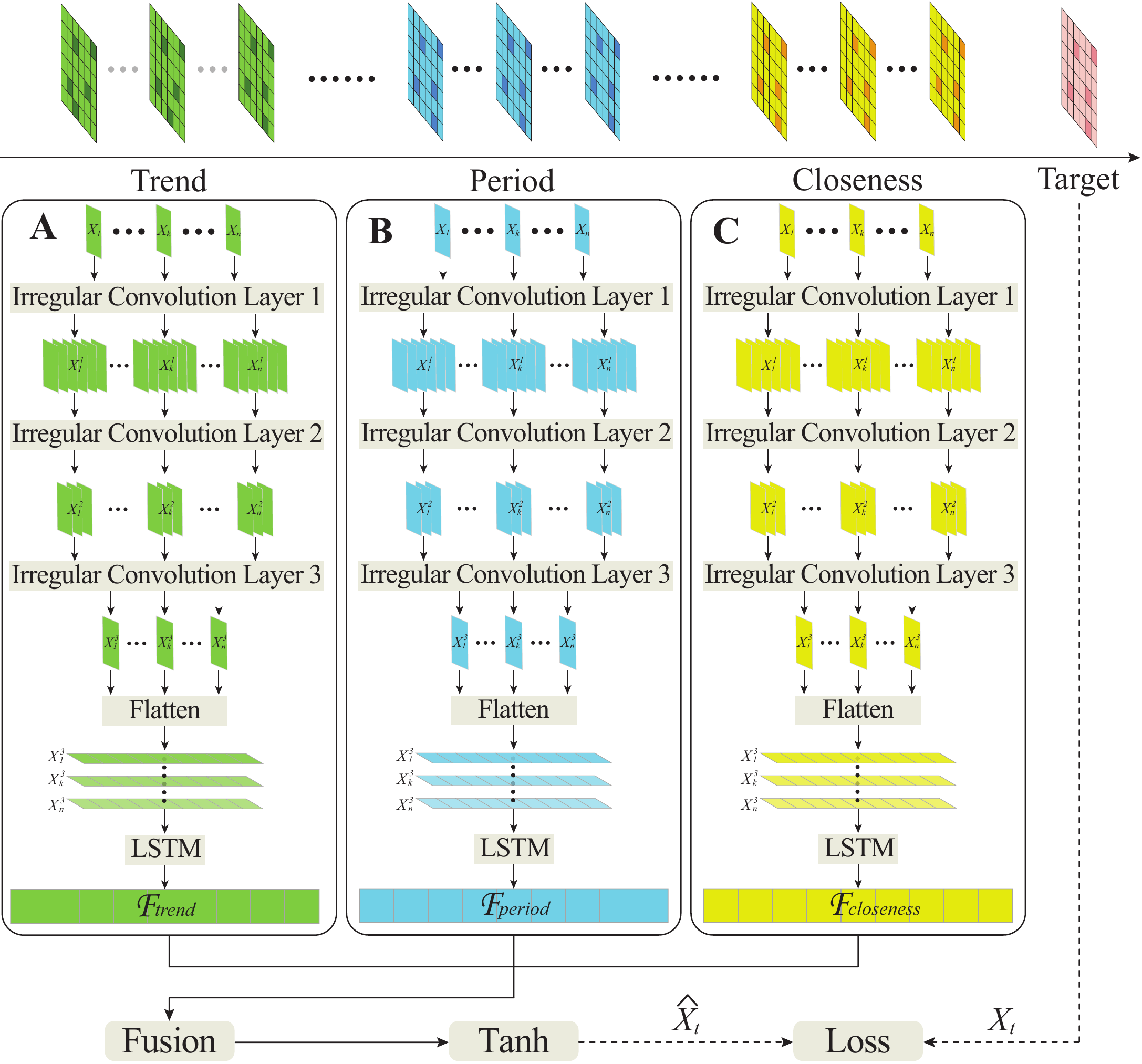}
	\caption{Overall architecture of the proposed model.}
	\label{fig:figure_2}
\end{figure}

Fig.~\ref{fig:figure_2} illustrates the overall architecture of the proposed model. The model consists of three separate modules with the same structure. Each module takes a specific set of historical observations (bike-sharing demand) as input. Instead of using all the historical observations to train the model, we identify key periods with different levels of recency to the target period (for which the prediction is made) and feed them into the three modules. The approach effectively reduces the time complexity of the training model by also mitigating the negative effect of redundant information in historical data. This practice has proved to achieve better performance than models trained using full historical observations \cite{zhang2017deep, ren2020hybrid, li2021short, wu2018hybrid}. The definitions of the three key periods (\textit{trend}, \textit{period} and \textit{closeness}) will be elaborated in Section~\ref{periods}.
As shown in Fig.~\ref{fig:figure_2}, each module adopts three layers of irregular convolutional architecture to capture the characteristics of bicycle demand among urban areas. The vector sequence formed by flattening the output of the irregular convolution is used as the input to the LSTM model\footnote{The description of the LSTM model is provided in \ref{appendix}.} to extract the temporal information in the sequence. The outputs of three hybrid modules are fed into a feature fusion layer. The output of the feature fusion layer is activated by a non-linear function generating the predicted value. The predicted value with its corresponding actual usage value participates loss estimation and backpropagation to update parameters in the model. In the following section, we formally introduce the architecture of irregular convolutional network.  

\subsection{Irregular convolutional neural network}
\begin{figure}[htbp]
\centering
	\includegraphics[width=6.5in]{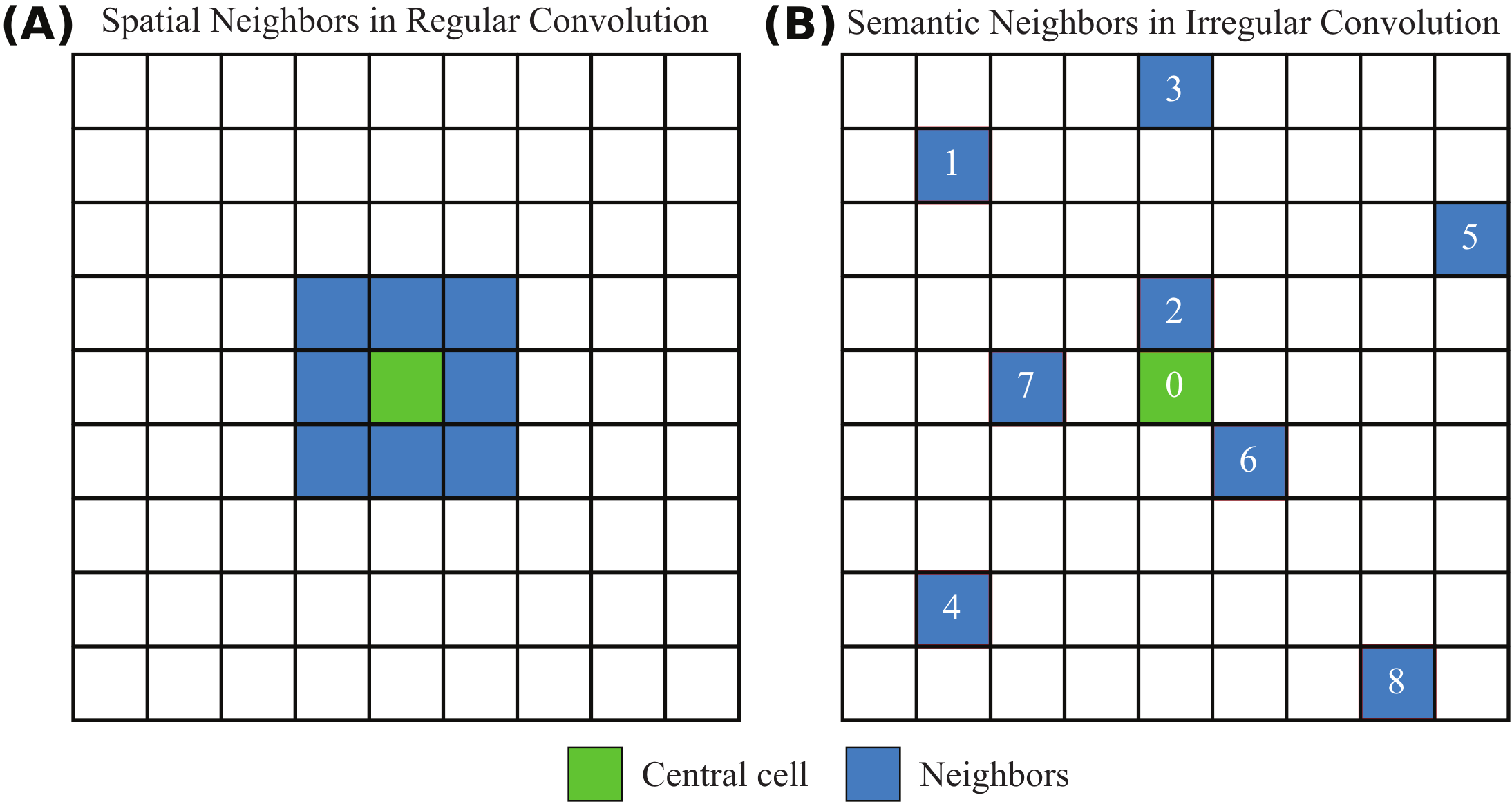}
	\caption{Illustration of traditional and irregular convolution with a kernel size of 9: \textbf{(A)} spatial neighbors are adjacent to the central cell in the regular convolution; \textbf{(B)} in the irregular convolution, semantic neighbors are identified based on the top 8 cells with the highest similarity of temporal bicycle usage patterns to the central cell.}
	\label{fig:two_ways_conv}
\end{figure}

The major difference between irregular convolution and traditional convolution lies in the cells involved in the convolutional operation. Generally, the number of cells involved in the convolution is known as the convolutional kernel size. Among cells corresponding to each kernel, the cell being forecast is called the central cell, and the other cells involved in convolution are called neighbors. Taking the convolutional kernel size of nine as an example, the neighbors involved in traditional convolution are spatially adjacent to the central cell, as shown in Fig.~\ref{fig:two_ways_conv}A. In contrast, the neighbors can be located anywhere in the study area for irregular convolution (Fig.~\ref{fig:two_ways_conv}B). In this study, for each central cell, we identify the top eight cells which show similar temporal bicycle usage patterns observed from the historical observation data. We call these cells as \textit{semantic neighbors}. Compared to the traditional convolution, irregular convolution is more flexible to exploit cells with similar temporal usage patterns to the central cell. 

This study uses two metrics, namely, the Pearson Correlation Coefficient~\cite{benesty2009pearson, kristoufek2014measuring} and Dynamic Time Warping (DTW)~\cite{muller2007dynamic, taylor2015method}, to quantify the similarity of temporal bicycle usage patterns between the cells. Because these two metrics measure temporal similarity from different perspectives, we aim to assess which metrics tends to result into a better prediction accuracy.

\begin{figure}[htbp]
\centering
	\includegraphics[width=6.2in]{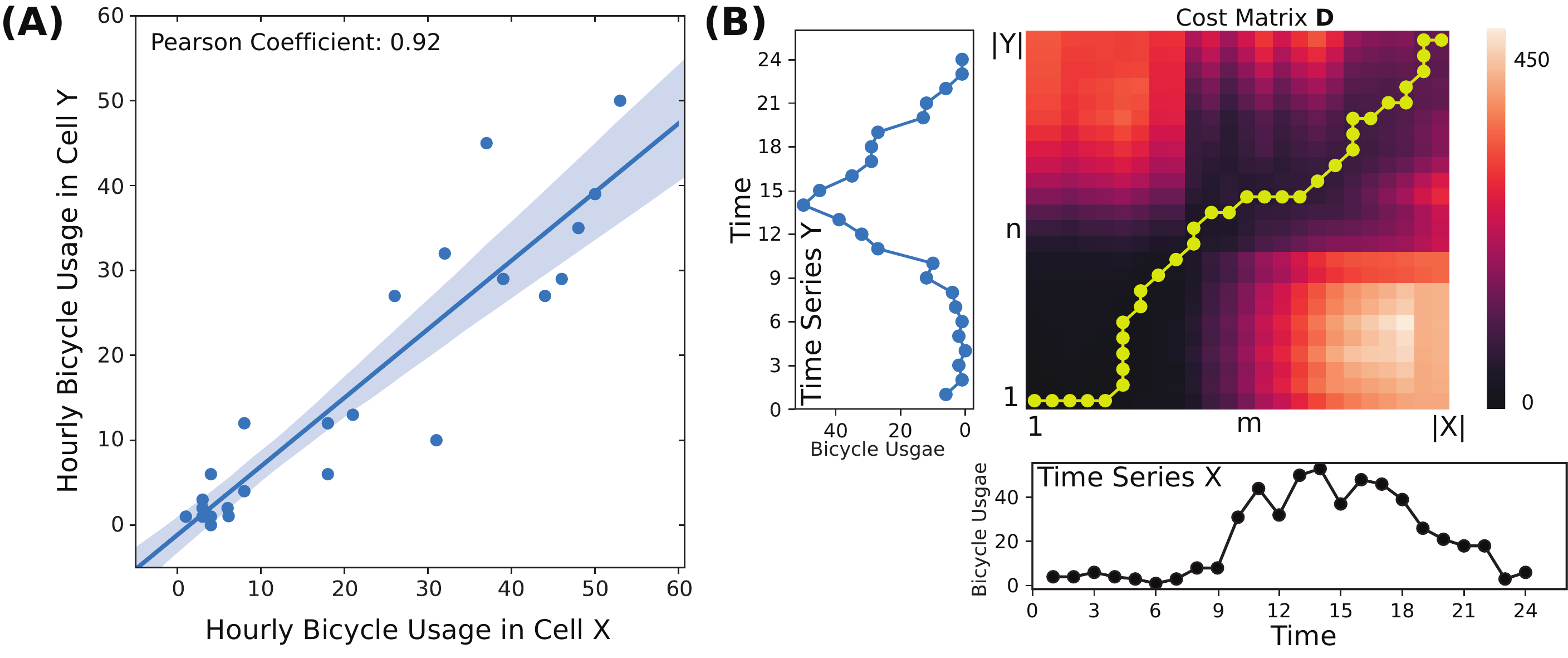}
	\caption{Two similarity metrics adopted in this study: \textbf{(A)} Pearson correlation coefficient; \textbf{(B)} Dynamic time warping (DTW).} 
	\label{fig:figure3_temp}
\end{figure}

The Pearson correlation coefficient measures the strength of a linear association between two sequences. The coefficient is the ratio between covariance of two variables and the product of their standard deviations. The two sequences are more positively correlated when the ratio is closer to 1. Eq.~\ref{equ:pearson} shows the calculation of the Pearson correlation coefficient. Fig.~\ref{fig:figure3_temp}A shows an example of the Pearson correlation coefficient of two bicycle usages sequences within 24 hours. 
\begin{equation}
r_{xy} = \frac{\sum_{i=1}^{n}{(X_i-\overline{X})(Y_i-\overline{Y})}}{\sqrt{\sum_{i=1}^{n}{(X_i-\overline{X})}^2\sum_{i=1}^{n}{(Y_i-\overline{Y})}^2}}
\label{equ:pearson}
\end{equation}
where $\overline{X}$ and $\overline{Y}$ denote the means of sequence $X$ and $Y$, respectively. $n$ denotes the length of sequence. 

For DTW, given two bicycle usage sequence $X={x_1, x_2,\cdots, x_i, \cdots, x_{|X|}}$ and $Y=y_1, y_2,\cdots\!,$ $ y_j, \cdots,  y_{|Y|}$, the optimization objective is to find a shortest warp distance between two sequences $dist(W)$ (typically Euclidean distance), as shown from Eq.~\ref{equ:dtw_1} to Eq.~\ref{equ:dtw_3}:

\begin{equation}
\begin{aligned}
& dist(W) = minimum(\sum_{k=1}^{k=K}{distance(w_{ki}, w_{kj})})\\
\end{aligned}
\label{equ:dtw_1}
\end{equation}
\begin{equation}
\begin{aligned}
& subject \ to:\\
& W=w_1, w_2, \cdots, w_K, \ max(|X|,|Y|)\leq K \textless |X|+|Y| \\
\end{aligned}
\label{equ:dtw_2}
\end{equation}
\begin{equation}
\begin{aligned}
& w_k=(i,j), w_{k+1}=(i^{\prime},j^{\prime}), \ i \leq i^{\prime} \leq i+1, j \leq j^{\prime} \leq j+1 \\
\end{aligned}
\label{equ:dtw_3}
\end{equation}
where $K$ is the length of warp path, $w_k$ denotes the $k^{th}$ element of the warp path, $i$ and $j$ represent the index of a certain record in two sequences $X$ and $Y$, respectively. $distance(w_{ki}, w_{kj})$ is the distance between two records (the former from $X$ and the latter from $Y$) in the $k^{th}$ element of the warp path. Thus, a cost matrix $D$ is constructed based on the distance from any record in one of sequences ($X/Y$) to any record in another sequence ($Y/X$). DTW adopts a greedy search from $D(|X|,|Y|)$ to $D(1,1)$ to find the minimum distance between two sequences. In this study, we use historical observations of each cell during the whole training period to estimate the similarity. The length of the evaluated sequences are the same for two mentioned metrics ($n=|X|=|Y|$). Fig.~\ref{fig:figure3_temp}B shows an example of finding the shortest-distance warp path based on the cost matrix $D$ for two bicycle usage sequences.  

The irregular convolutional computation is analogous to the traditional convolution:

\begin{equation}
y(i,j) = b(i,j) + \sum_{c=1}^{C_{in}}{\sum_{s=1}^{S}{x^s_c(i,j)w^s_c}} 
\label{equ:conv}
\end{equation}
where $C_{in}$ denotes the number of channels in input $x(i,\,j)$; $S$ denotes the size of convolutional kernel; $x^s_c(i,j)$ represents the semantic neighbor $s$ associated with central cell $x(i,j)$ in the channel $c$; $w^s_c$ denotes the weight in the convolutional kernel corresponding to the semantic neighbor $x^s_c(i,j)$; $b(i,j)$ denotes the learnable bias. 

\subsection{Definitions of Trend, Period and Closeness}\label{periods}
As mentioned in~\ref{architecture}, given a target period (for which the prediction is made), we identify three key periods from historical observations according to the levels of recency to the target period. We name them as $Closeness$, $Period$, and $Trend$ (Fig.~\ref{fig:figure_2}). The purpose of 
selecting observations from these periods as inputs is to reduce the negative impact of redundant information in the whole training data on model performance while lowering the training time complexity. For example, the usage patterns of shared bicycles during peak hours may be similar during weekdays, and bicycle usage might be associated with that at the same time in the previous week. This strategy of identifying key periods as training input has proved to achieve better performance than models trained using full historical observations~\cite{zhang2017deep, ren2020hybrid, li2021short, wu2018hybrid}. In this study, the definitions of such three key periods are given in Eq.~\ref{equ:hitorical periods_1} to Eq.~\ref{equ:hitorical periods_3}.

\begin{equation}
\begin{aligned}
\mathcal{X}_{t}^{closeness} = \{X_{t-l_c}, X_{t-(l_c-1)}, \cdots, X_{t-1}\}\\
\end{aligned}
\label{equ:hitorical periods_1}
\end{equation}
\begin{equation}
\begin{aligned}
\mathcal{X}_{t}^{period} = \{X_{t-24*l_p}, X_{t-24*(l_p-1)}, \cdots, X_{t-24}\}\\
\end{aligned}
\label{equ:hitorical periods_2}
\end{equation}
\begin{equation}
\begin{aligned}
\mathcal{X}_{t}^{trend} = \{X_{t-7*24*l_q}, X_{t-7*24*(l_q-1)}, \cdots, X_{t-7*24}\}
\end{aligned}
\label{equ:hitorical periods_3}
\end{equation}

We select $l_c$ as 24 to capture historical shared bicycle usage data in the past 24 hours. The value of $l_p$ is set as 7 to select the usage data at the same time for each day in the past week. $l_q$ is chosen as 2 to provide historical information at the same time in the past week and the week before the past. The time complexity of training a model using the above mentioned key periods will be lower than that of training with a lengthy period of data.

The outputs of three separate modules are fused for the final forecast. We adopt the weighted element-wise addition method to merge the three spatial-temporal features, including $\mathcal{F}_{trend}$, $\mathcal{F}_{period}$ and $\mathcal{F}_{closeness}$. Then, the feature map is activated by a $tanh$ function to generate the prediction values $\hat{X}_t$ that participates in the loss and backpropagation with the actual bicycle usage $X_t$. The computation of the prediction values $\hat{X}_t$ is shown Eq.~\ref{equ:fusion}.

\begin{equation}
\hat{X}_t = tanh(\mathcal{F})= tanh(W_t \circ \mathcal{F}_{trend} + W_p \circ \mathcal{F}_{period} + W_c \circ \mathcal{F}_{closeness})
\label{equ:fusion}
\end{equation}
where $W_t$, $W_p$ and $W_c$ denote the learnable parametric vectors with the same shape size of the corresponding feature, $\mathcal{F}$ is the feature map after fusion, and $tanh$ denotes the activation function.

\subsection{Hyperparameter settings and benchmark models}
Since we adopt two metrics to quantify the similarity of temporal usage patterns, the performance of two variants of IrConv+LSTM is evaluated in this study. We name them as IrConv+LSTM:P and IrConv+LSTM:D, respectively. The hyperparameters for both variants are the same and refer to several existing studies~\cite{liu2018short, yang2019mf, zhang2018citywide}. Three irregular convolutional layers in each separate module operation (shown in Fig.~\ref{fig:figure_2}) are adopted with 32 filters, 16 filters, and 1 filter, respectively. The last filter in the irregular architecture is to aggregate the high-dimensional information into one channel. The convolutional kernel size for two variants of our model is set as nine.

Four baseline models are adopted for performance comparison with two IrConv+LSTM variants, including one parametric model (ARIMA) and three deep learning models (LSTM, STRN, and CNN+LSTM): 

\begin{description}[itemindent=3pt, font=$\bullet$~\normalfont\scshape]
\item [ARIMA:] Auto-Regressive Integrated Moving Average model is a parametric model widely used for time series forecasting. ARIMA is a combination of the differenced autoregressive model (AR) with the moving average model (MA)~\cite{kumar1999autoregressive}. ARIMA can handle non-stationary sequences by replacing the data values with the difference between their values and the previous values to obtain stationary sequences. Thus, ARIMA is widely adopted to predict traffic status that is dynamically changed over time, such as traffic accidents, traffic speed, and traffic volume~\cite{avuglah2014application, billings2006application, lee1999application}. 

\item [LSTM:] Long Short-Term Memory is a widely used deep learning architecture in the field of traffic prediction~\cite{zhao2017lstm,fu2016using,shin2020prediction}. As a variant of Recurrent Neural Network (RNN), one of the advantages of LSTM is that it can capture the information for both long and short periods of time by introducing gate theory. LSTM is effective for processing long sequences of input data because it tackles the gradient vanishing and explosion problems that can be encountered when training traditional RNN models. 

\item [STRN:] Spatial-Temporal Residual Network is a hybrid deep learning prediction model initially proposed to forecast short-term pedestrian volume~\cite{zhang2017deep}. It consists of three Residual Convolutional Neural Networks (ResNet). The spatial features captured by ResNets from multiple fragments of historical data are fused together for the final prediction. The strategy of adopting fragments of historical data to train deep learning prediction models has been referenced in many studies~\cite{liu2019deeppf,ren2020hybrid,li2021short,du2019deep}. 

\item [CNN+LSTM:] CNN+LSTM is a hybrid deep learning model for spatial-temporal prediction. This model couples a traditional convolution and an LSTM model to extract historical spatial-temporal information for the prediction. It has been adopted for traffic prediction by several studies~\cite{kim2019predicting,cao2020cnn}. The difference between this model and our proposed model is that CNN+LSTM adopts traditional convolution involving spatial neighbors. 
\end{description}

To make the prediction results of IrConv+LSTM and CNN+LSTM comparable, we set similar hyperparameters and structures for both of them. Specifically, three hybrid modules in CNN+LSTM extract spatial-temporal information  from their respective key historical periods for modeling. The definitions of such three key periods in CNN+LSTM is the same as them in our model. Each separate module also contains three traditional convolution layers with 32 filters in the $1^{st}$ layer, 16 filters in the $2^{nd}$ layer, 1 filter in the $3^{rd}$ layer. The convolutional kernel size adopted in CNN+LSTM is also set as nine. The hyperparameter settings of STRN are referred to the settings in article~\cite{zhang2017deep}. The parameters of all benchmarks have been calibrated to achieve optimal prediction results. 

We use the first 80\% of the hourly usage data as the training data and the last 20\% data to validate the performance of models in each city, as shown in Table~\ref{table:study_area}. The loss function of all deep learning models in this study is the MSELoss function. Eq.~\ref{equ:MSE} defines the MSELoss function that is used to measure the errors between the prediction value and the ground truth~\cite{wang2016training}. The optimization algorithm for updating parameters in deep learning models is important for backpropagation. This study employs RMSProp (Root Mean Square Prop) as the optimization algorithm across all deep learning models~\cite{graves2013generating}. Moreover, three indicators are employed to evaluate the performance of models, including Mean Absolute Percentage Error (MAPE), Root Mean Square Error (RMSE), and Mean Absolute Error (MAE)~\cite{zhang2017deep,ren2020hybrid,zhang2020novel}. The definitions of them are shown from Eq.~\ref{con:MAPE} to Eq.~\ref{con:MAE}.
\begin{equation}
MSE = \frac{1}{n}\sum_{i=1}^n(\hat{y}_i-y_i)^2
\label{equ:MSE}
\end{equation}
\begin{equation}
MAPE =  \frac{100\%}{n}  \sum_{i=1}^{n} |\frac{\hat{y}_i-y_i}{y_i}|
\label{con:MAPE}
\end{equation}

\begin{equation}
RMSE = \sqrt{\frac{1}{n}\sum_{i=1}^n(\hat{y}_i-y_i)^2}
\label{con:RMSE}
\end{equation}

\begin{equation}
MAE = \frac{1}{n} \sum_{i=1}^{n}\left | \hat{y}_i -y_i \right |
\label{con:MAE}
\end{equation}
 
Here $\hat{y}_i$ denotes the prediction result and $y_i$ denotes the actual value of bicycle usage. All deep learning frameworks are constructed on the Pytorch platform~\cite{NEURIPS2019_9015}. Also, the models are built on a server with NVIDIA Tesla V100 and a workstation with NVIDIA RTX 2070 Super Graphics Card.

\section{Analysis Results}\label{results}
\subsection{Overall accuracy of the proposed model and benchmark models}\label{overall}
In this section, we compare the overall accuracy between our proposed model and the benchmark models. Table~\ref{table:overall} shows the overall accuracy of two variants of our proposed model and other benchmark models. Given a specific indicator, the model with the best performance is marked with $*$ in Table~\ref{table:overall}.
\begin{table}[htbp]
\caption{Overall accuracy of all models across five cities}
\label{table:overall}
\centering
\setlength\extrarowheight{1.1pt}
\resizebox{\textwidth}{!}{
\begin{tabular}{cccccccc} 
\bottomrule
\textbf{City}                                                                                                                                                         & \textbf{Index} & \textbf{ARIMA} & \textbf{LSTM} & \textbf{STRN} & \textbf{\begin{tabular}[c]{@{}c@{}}CNN+\\LSTM\end{tabular}} & \textbf{\begin{tabular}[c]{@{}c@{}}IrConv-\\ LSTM:P\end{tabular}} & \textbf{\begin{tabular}[c]{@{}c@{}}IrConv-\\ LSTM:D\end{tabular}}  \\ 
\bottomrule
\multirow{3}{*}{\textbf{Singapore }}                                                                                                                                  & \textbf{MAPE}  & 0.8488         & 0.6715        & 0.7026      & 0.6696      & $0.5617^*$            & 0.5638             \\
                                                                                                                                                                      & \textbf{MAE}   & 2.6267         & 2.2411        & 2.2488      & 2.0971     & 1.9727            & $1.9655^*$             \\
                                                                                                                                                                      & \textbf{RMSE}  & 4.0656         & 3.4403        & 3.3602      & 3.2243     & 3.0911            & $3.0764^*$             \\ 
\midrule
\multirow{3}{*}{\textbf{Chicago }}                                                                                                                                    & \textbf{MAPE}  & 1.3000         & 0.8615        & 0.7131      & 0.7584     & 0.7240            & $0.6028^*$             \\
                                                                                                                                                                      & \textbf{MAE}   & 4.6156         & 3.6955        & 2.5619      & 2.8373     & 2.4036            & $2.2047^*$             \\
                                                                                                                                                                      & \textbf{RMSE}  & 12.6296        & 10.1246       & 4.7583      & 6.5034      & 4.7802            & $4.3356^*$             \\ 
\midrule
\multirow{3}{*}{\begin{tabular}[c]{@{}c@{}}\textcolor[rgb]{0.125,0.129,0.141}{\textbf{Washington, }}\\\textcolor[rgb]{0.125,0.129,0.141}{\textbf{D.C.}}\end{tabular}} & \textbf{MAPE}  & 0.9672         & 0.7133        & 0.6858      & 0.6492      & $0.5599^*$            & 0.5675             \\
                                                                                                                                                                      & \textbf{MAE}   & 3.7649         & 2.3535        & 2.3327      & 2.4026     & 1.9758            & $1.9697^*$             \\
                                                                                                                                                                      & \textbf{RMSE}  & 7.2917         & 4.0227        & 4.0657      & 4.0718     & 3.3669            & $3.3590^*$             \\ 
\midrule
\multirow{3}{*}{\textbf{New York }}                                                                                                                                   & \textbf{MAPE}  & 2.9407         & 0.7479        & 0.7770      & 0.7153      & 0.6243            & $0.6201^*$             \\
                                                                                                                                                                      & \textbf{MAE}   & 16.9310        & 7.5721        & 7.2278      & 6.1815     & 5.9865            & $5.7769^*$             \\
                                                                                                                                                                      & \textbf{RMSE}  & 31.7418        & 15.3383       & 13.7237     & 11.3954     & 11.3385           & $10.8141^*$            \\ 
\midrule
\multirow{3}{*}{\textbf{London }}                                                                                                                                     & \textbf{MAPE}  & 1.7924         & 0.7483        & 0.6753      & 0.6862     & 0.5852            & $0.5523^*$             \\
                                                                                                                                                                      & \textbf{MAE}   & 8.7771         & 4.8125        & 4.2348      & 4.0423     & 3.7829            & $3.5785^*$             \\
                                                                                                                                                                      & \textbf{RMSE}  & 17.8831        & 9.4463        & 7.1533      & 7.2279     & 6.6326            & $6.2960^*$             \\
\bottomrule
\end{tabular}}
\end{table}

Generally, our proposed model achieves better performance than other baseline models across five cities based on the prediction accuracy. As a typical parametric model, ARIMA is hard to process the non-stationary sequence and cannot leverage any spatial dependency of bike-sharing usage. Thus, it achieves the lowest accuracy of forecasting bicycle demand across five cities. Unlike ARIMA, LSTM model is effective to extract temporal information of bicycle usage in each cell from non-stationary sequence by using the mechanism of deep learning technology. Hence, the results of LSTM are better than that of ARIMA. However, LSTM still cannot utilize the spatial dependency of bicycle usage among cells for prediction. Based on the results, the hybrid deep learning model that couples convolution architectures and LSTM model to extract spatial-temporal features of bicycle usage generally outperforms LSTM in all cities. Moreover, STRN also performs better than LSTM. Although STRN and CNN+LSTM achieve good prediction results, the performance of our proposed model is still better than them in all indicators. Notably, our proposed model achieves an improvement of MAPE from 8\% (in Washington, D.C.) to 12\% (in London), compared to the model with the best performance in four benchmarks.

Compared to CNN+LSTM, our proposed model only replaces the spatial neighbors involved in regular convolution with the semantic neighbors adopted in irregular convolution. However, the prediction accuracy of our model is much higher than CNN+LSTM across five cities. Specifically, among five cities, our approach achieves an improvement of MAPE by 8\%-15\% compared to CNN+LSTM. Such results imply that the semantic neighbors are more effective than spatial neighbors for predicting bike-sharing usage. The semantic neighbors are identified according to the similarity of temporal bicycle usage to their corresponding central cells. Although the semantic neighbors are not always spatially adjacent to their central cells, they can provide essential information for spatial-temporal modeling than the spatial neighbors. Therefore, the prediction results suggest that involving semantic neighbors in the irregular convolution is a good strategy for predicting shared-bicycle usage across all study areas.
 
We also find that the performance of the variant with DTW metric is better than the variant with Pearson measure in most cities. The variant with Pearson correlation coefficient only performs slightly better in two cities (Singapore and Washington D.C.) from the perspective of MAPE. Based on the characteristics of Pearson correlation coefficient, the usage variations of two measured sequences are simply reflected by their respective standard deviations, and the temporal offsets between such two sequences cannot be quantified by the coefficient. In contrast, DTW metric is effective to measure the similarity of usage variations between two sequences by calculating the distance of each record in one sequence to all other observations in the other. Also, the shortest warping distance between the sequences searched by DTW metric reflects the similarity without the effects of temporal offsets between the sequences. In other words, for two sequences that are not synchronized, DTW metric is able to quantify their similarity more precisely than Pearson correlation coefficient. Especially for quantifying the similarity of temporal bicycle usage patterns, DTW metric, which considers the temporal offsets, better reflects the similarity of travel behavior among areas. In sum, our proposed model achieves the best overall performance compared to other baseline models. Also, DTW metric is an effective way to select semantic neighbors involved in the irregular convolution.

\begin{figure}[!h]
\centering
	\includegraphics[width=6in]{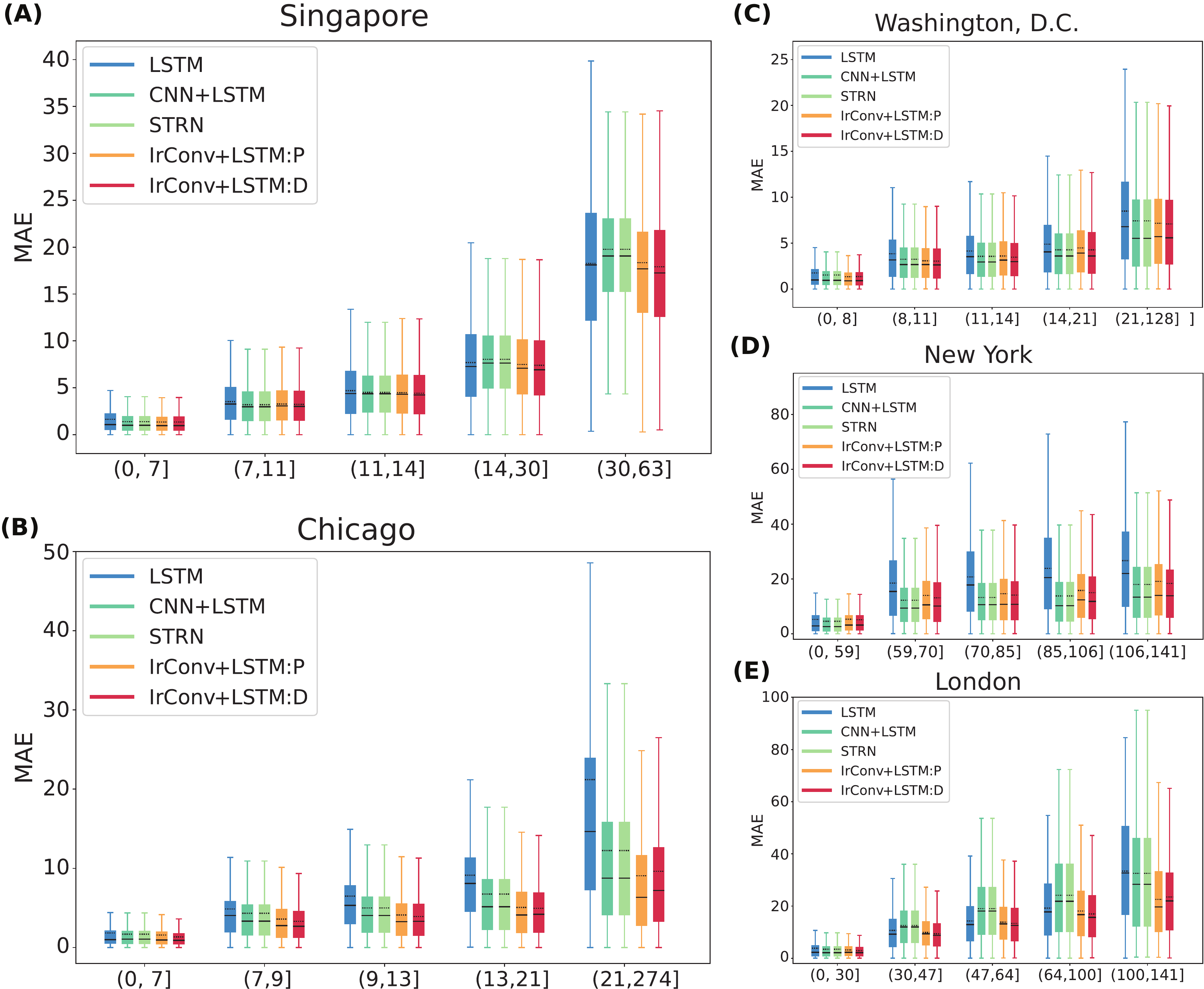}
	\caption{Performance of models in cells with various levels of bicycle usage. The cells in each city are divided into five quantiles according to their respective hourly usage, in order from low to high usage shown in horizontal axes.}
	\label{fig:figure_4}
\end{figure}

\subsection{Performance of models over cells with varying levels of bicycle usage}
When operating bike-sharing systems, one of the important tasks is to satisfy users' travel needs. However, bike-sharing demand is not evenly distributed in urban areas. The number of areas with high user demand is relatively small but most users intend to ride bicycles in such areas. In contrast, areas with low demand require fewer bicycles, but they are widely distributed in the city. Therefore, there is a trade-off when allocating bicycles across urban areas. If the travel demand in areas with different usage can be precisely predicted, it is helpful for the deployment of bicycles. 

We separate the cells into five quantiles in each city according to their respective hourly usage to assess the models' performance over these quantiles. Fig.~\ref{fig:figure_4} shows the MAE distributions of each quantile of bicycle usage in the five cities. We find that the performance of our model is better than baseline models across five quantiles of bicycle usage in most cities.  Particularly, for these high-demand cells, the prediction accuracy of our proposed model is higher than other benchmarks. For example, the average MAE of our approach is less than 17 bicycles in Singapore, lower than that of the benchmarks. Also, the maximum MAE in Chicago is close to 10 bicycles, much smaller than other benchmarks. However, in Washington, D.C., and New York, the average MAE of our approach is similar to that of other baseline models. This fact indicates that our approach has a similar prediction ability to the other models in such two cities with large bicycle usage. Additionally, our proposed model achieves better performance on the other quantiles of bicycle usage. Even in low-demand areas in most cities, our proposed model's average MAE is lower than the other baseline models. To sum up, IrConv+LSTM has achieved better performance over areas with varying levels of bicycle usage. 
\begin{figure}[!h]
\centering
	\includegraphics[width=6in]{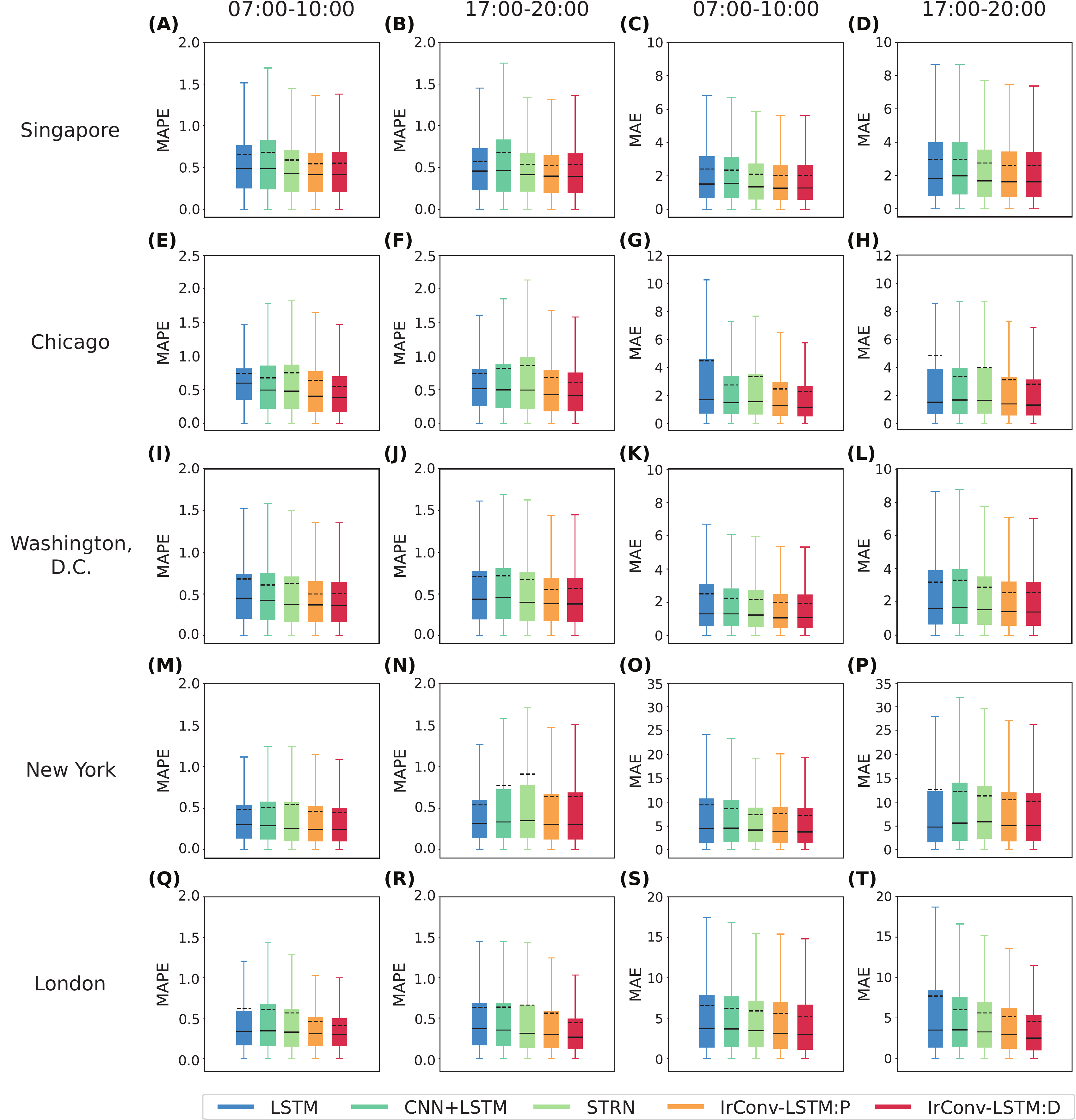}
	\caption{Performance of models during peak hours: \textbf{(A-D)} Singapore; \textbf{(E-H)} Chicago; \textbf{(I-L)} Washington D.C.; \textbf{(M-P)} New York; and \textbf{(Q-T)} London. }
	\label{fig:figure_6}
\end{figure}

\subsection{Performance of models during peak hours}
Another important task in operating bike-sharing systems is meeting the users' needs during the morning and evening commuting hours. For example, in some cities, there are many users who prefer to ride bicycles to address first- and/or last-mile problems during the morning and evening peak hours~\cite{xu2019unravel,midgley2011bicycle}. The operators need to allocate enough available bicycles for users during such peak hours. Accurate bicycle demand forecasting can help them develop proper scheduling plans to satisfy users with as little cost as possible. 

We select the morning peak (7:00-10:00 AM) and evening peak (5:00-8:00 PM) observations from the validation datasets to assess the models' performance in each city. Fig.~\ref{fig:figure_6} represents the prediction accuracy of each model during peak hours in five cities based on the two indicators (MAPE and MAE). The prediction error of our proposed model is the lowest across five cities during both peak hours. Specifically, in London, the MAE of IrConv+LSTM:D is smaller than five during the evening period, better than the benchmarks. In Chicago, the MAPE of our approach is close to 50\% during both peak periods, which is smaller than other baseline models. Although IrConv+LSTM performs similarly to CNN+LSTM in Singapore during both peak hours, the performance of IrConv+LSTM is stable and better across all five cities. Comparing the two variants of our proposed model, the performance of the variant adopted DTW metric is more robust in most cities than the variant with the Pearson measure. To sum up, our model outperforms the benchmarks during peak hours in five cities. 

\subsection{Comparative analysis between semantic and spatial neighbors}\label{Comparative}
As mentioned in Section~\ref{overall}, our approach involving semantic neighbors in irregular convolution has a notable improvement in model performance compared to the benchmark (CNN+LSTM) incorporating spatial neighbors in traditional convolution. Here we perform an additional analysis to gain insights into the spatial relationship between cells' semantic and spatial neighbors.

\begin{figure}[htbp]
\centering
	\includegraphics[width=6.5in]{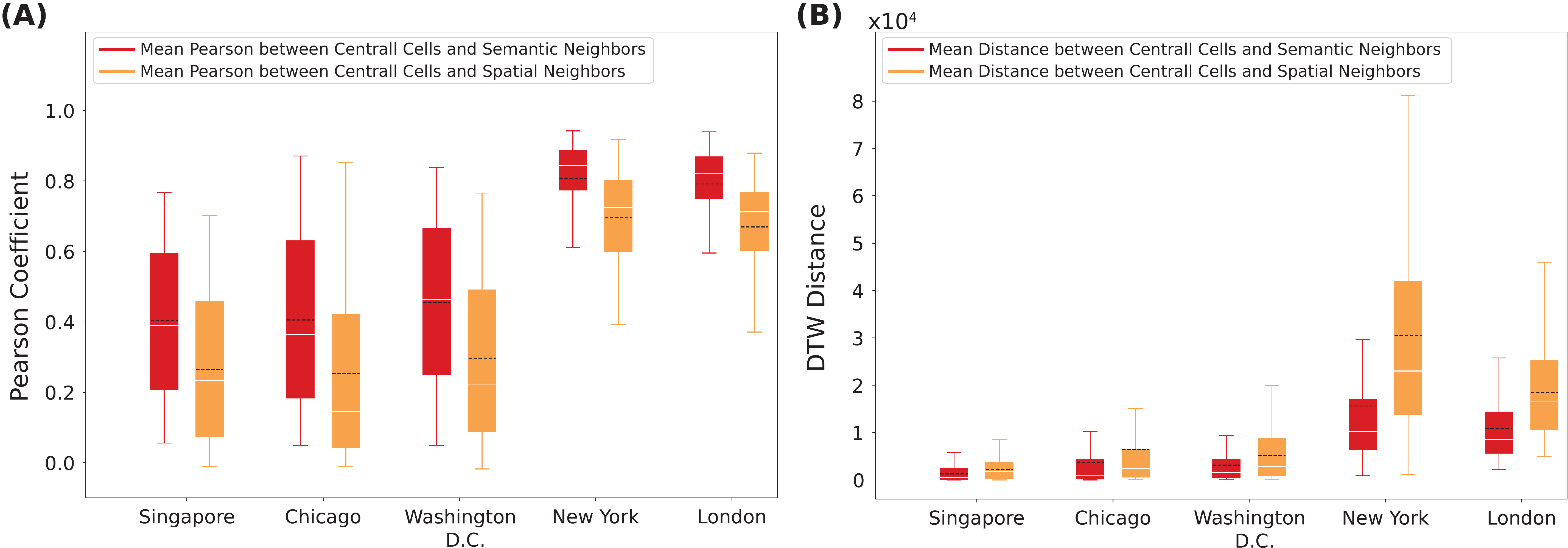}
	\caption{\textbf{(A)} Distribution of average Pearson correlation coefficients between the central cell and spatial neighbors vs. semantic neighbors; \textbf{(B)} Distribution of average DTW distance between the central cell and spatial neighbors vs. semantic neighbors.}
	\label{fig:mean_weights}
\end{figure}

First, we analyze the difference in temporal similarity between the central cell and spatial neighbors vs. the central cell and semantic neighbors. As shown in Fig.~\ref{fig:mean_weights}, the similarity of temporal usage patterns between central cells and their semantic neighbors is generally higher than that between central cells and their spatial neighbors. Combined with the prediction accuracy of our model and CNN+LSTM, the similarity of semantic neighbors is more effective in enhancing the accuracy of deep learning models. However, the traditional convolution fails to leverage such similarity due to its structural limitations. Moreover, the similarity differences between the two types of neighbors imply that directly applying traditional convolution involving spatial neighbors for shared bicycle usage prediction might not always achieve desirable performance. In the five study sites evaluated in this research, CNN+LSTM based on regular convolution architecture performs worse than our proposed model.

\begin{figure}[htbp]
\centering
	\includegraphics[width=6.5in]{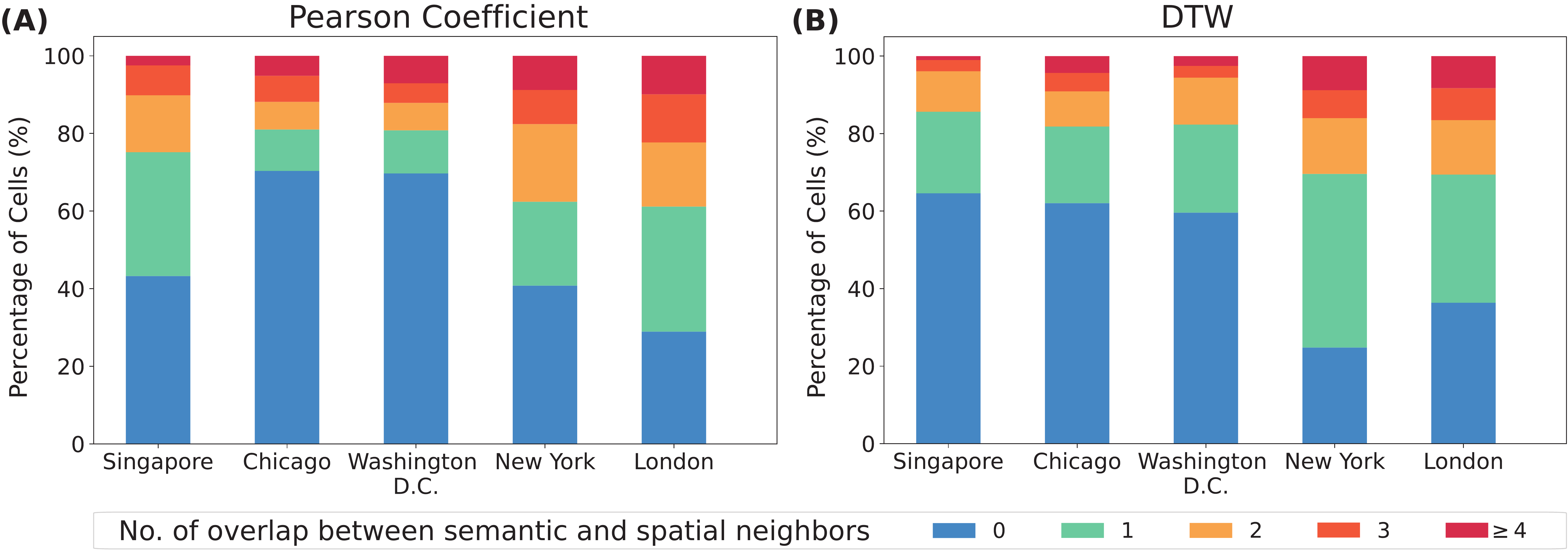}
	\caption{Number of overlapping cells between a central's spatial and semantic neighbors: \textbf{(A)} results based on Pearson correlation coefficient; \textbf{(B)} results based on Dynamic Time Warping (DTW).}
	\label{fig:overlaps}
\end{figure}

From a spatial point of view, we evaluate the extent to which semantic neighbors overlap with spatial neighbors. Fig. \ref{fig:overlaps} shows the distribution of overlap cells between spatial and semantic neighbors in five cities. We find that more than 70\% of central cells have their spatial and semantic neighbors sharing less than two cells, which can be observed under both Pearson and DTW metrics. This fact indicates that the semantic neighbors are mainly distributed at non-spatially adjacent areas to the central cells, further illustrating the difference between semantic and spatial neighbors. In other words, some areas in a city, though far apart, could show relatively similar temporal bicycle usage patterns than spatially adjacent areas. This is potentially attributed to built environment characteristics, urban functions, and other factors that shape users' travel behavior across urban areas.

\section{Conclusion and Discussion}\label{concolusion}

This paper proposes a hybrid deep learning model coupling irregular convolution architectures and LSTM modules to predict bicycle usage demand across one dockless bike-sharing system in Singapore and four station-based systems in Chicago, Washington D.C., New York, and London. The irregular convolution is applied over semantic neighbors that refer to places with temporal usage patterns similar to those of the areas being forecast. To measure the cells' similarity and build the semantic neighbors, one variant of our model uses the Pearson correlation coefficient while the other adopts Dynamic Time Warping (DTW). Our proposed model and four benchmark models are evaluated in five cities. Based on the prediction results, the proposed model generally outperforms the benchmarks across five cities. The prediction accuracy of the proposed model is also higher than that of the benchmarks in areas with different bicycle usage levels and during peak hours. Comparing the two variants of our model, the performance of the variant with the DTW metric is better than that of the variant with the Pearson coefficient metric in most of the cities. 

We find that the semantic neighbors adopted in the irregular convolution are quite different from the spatial neighbors involved in the traditional convolution. Specifically, the similarity of temporal usage patterns between the central cells and their semantic neighbors is generally higher than that between the central cells and their spatial neighbors under both the Pearson and DTW metrics. Unlike the spatial neighbors, the semantic neighbors are mainly distributed in areas that are not spatially adjacent to their central cells. The model comparison suggests that relating areas that share similar temporal usage patterns through irregular convolution is helpful to improving the prediction accuracy. The findings also indicate that the neighbors involved in the convolution and the metrics for quantifying similarity (e.g., Pearson correlation vs. DTW) among urban areas tend to influence the performance of the prediction model. 

The study suggests that ``thinking beyond spatial neighbors'' can inspire new solutions that improve short-term travel demand prediction. The implications go beyond applications of bike-sharing systems. In general, a reliable travel demand or traffic prediction model requires a good understanding on the spatial-temporal dependency of historical travel demand. Although the subjects being studied could vary (e.g., vehicles, cyclist, pedestrians and goods), it is reasonable to assume that there exists spatial autocorrelation in travel patterns. This partially explains why deep learning architecture with typical CNN (i.e., by capturing spatial dependency) is helpful to improving travel demand prediction in different transportation applications. However, given that travel patterns are affected by spatial variations of built environment characteristics, urban functions and socioeconomic characteristics, areas close to each other might show different travel dynamics that are sometimes not correlated over time. Therefore, the typical CNN architecture can sometimes introduce ``noise'' into the prediction process. The irregular convolution with semantic neighbors is one example of many possible strategies to overcome this limitation.

In this study, the semantic neighbors are identified solely based on the temporal similarity in historical bicycle usage patterns. These areas with similar temporal travel patterns could signify potential similarities in built environment characteristics and other factors that shape cycling behaviors. However, these environmental and socioeconomic characteristics are not directly utilized in our proposed model. A possible direction for future research is to consider both travel patterns and environmental factors for defining similarities and identifying semantic neighbors. Combing these static (e.g., built environment characteristics) and dynamic features (i.e., bicycle usage patterns) may further improve the robustness of the prediction model.

\appendix
\section{The Structure of Long Short-Term Memory Network}\label{appendix}
\begin{figure}[htbp]
\centering
	\includegraphics[width=4in]{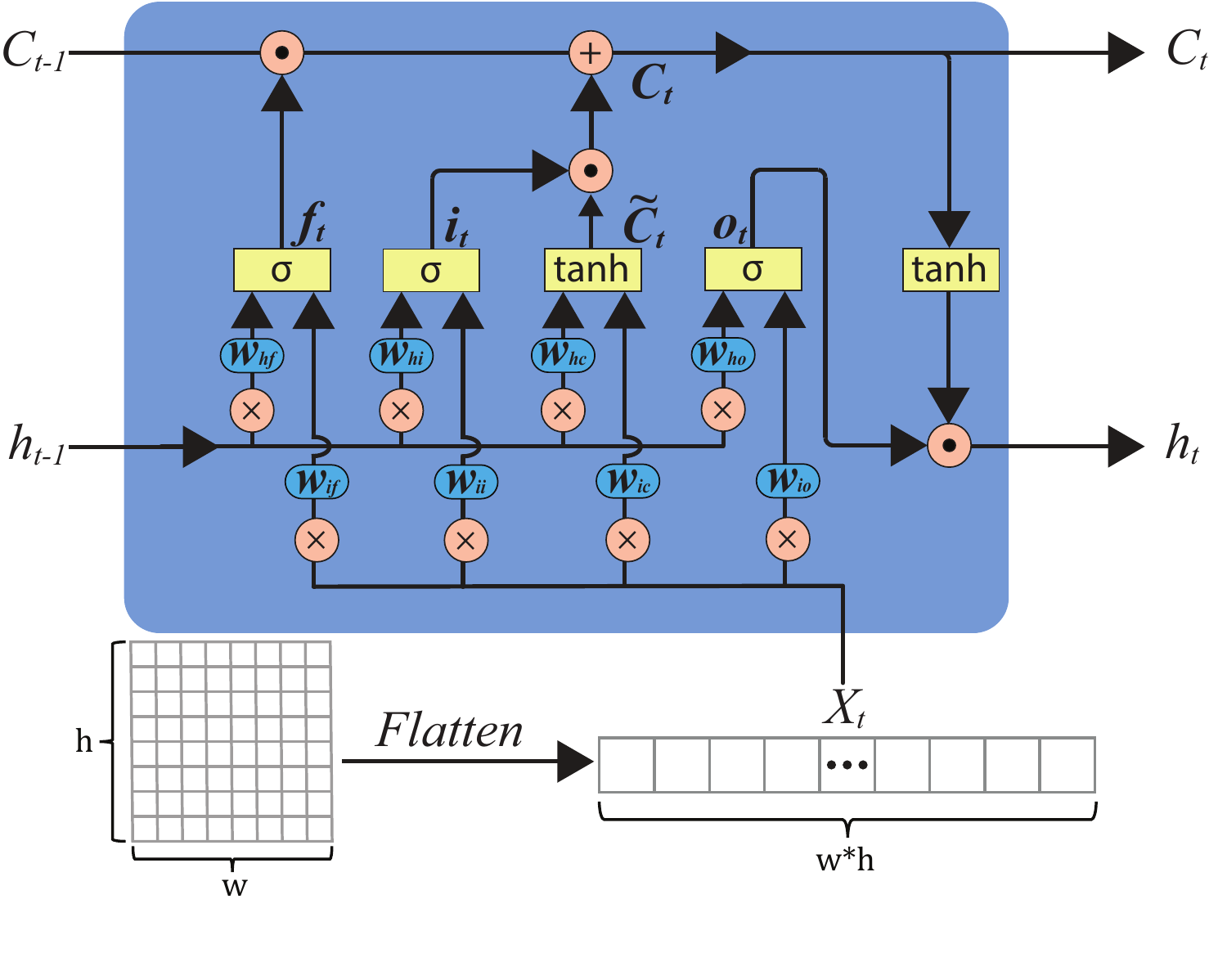}
	\caption{The structure of Long Short-Term Memory Network}
	\label{fig:lstm}
\end{figure}

We adopt the Long Short-Term Memory (LSTM) model to extract temporal information from the features captured by the irregular convolutional network~\cite{xiangxue2019data,fu2016using,du2019deep, hochreiter1997long}. The structure of LSTM is shown in Fig.~\ref{fig:lstm}. As a variant of Recurrent Neural Network, LSTM adopts the gate theory to control the information captured from both long and short periods of time for avoiding gradient descent or explosion. The computations of LSTM are shown from Eq.~\ref{equ:forget_gate} to Eq.~\ref{equ:ht}.
\begin{equation}
f_t = \sigma(W_{if}x_t+b_{if}+W_{hf}h_{t-1}+b_{hf})
\label{equ:forget_gate}
\end{equation}
\begin{equation}
i_t = \sigma(W_{ii}x_t+b_{ii}+W_{hi}h_{t-1}+b_{hi})
\label{equ:input_gate}
\end{equation}
\begin{equation}
\widetilde{C}_t = tanh(W_{ic}x_t+b_{ic}+W_{hc}h_{t-1}+b_{hc})
\label{equ:potensial_Ct}
\end{equation}
\begin{equation}
o_t = \sigma(W_{io}x_t+b_{io}+W_{ho}h_{t-1}+b_{ho})
\label{equ:O_gate}
\end{equation}
\begin{equation}
C_t = f_t \odot C_{t-1}+i_t \odot \widetilde{C}_t
\label{equ:Ct}
\end{equation}
\begin{equation}
h_t = o_t \odot tanh(C_t)
\label{equ:ht}
\end{equation}
Here $f_t$, $i_t$ and $o_t$ are the forget, input and output gates, respectively. $W_*$ and $b_*$ are respectively learnable weights and bias. $\odot$ is the Hadamard product, and $\sigma$ denotes the sigmoid function. Particularly, the input $x_t$ is the feature captured by irregular convolutional network at time $t$; $C_t$ denotes the cell state at time $t$ aggregating the information from forget, input gates, cell state and hidden state of the previous layer; $h_t$ represents the hidden state that incorporates the information from $C_t$ and the output gate. According to the equations of LSTM, the hidden state is essential for controlling short-term temporal information, while the cell state stores the long-term temporal information. Therefore, LSTM achieves good performance in prediction by capturing short- and long-term temporal information. Besides, LSTM only accepts vectors as the input in each time interval. As a result, the information captured by the irregular convolution are flattened as a vector then input to LSTM model (shown as Fig.~\ref{fig:lstm}).

\newpage

\bibliographystyle{elsarticle-num} 
\bibliography{literature}





\end{document}